\documentclass[letterpaper, 10 pt, conference]{ieeeconf}

\IEEEoverridecommandlockouts
\usepackage{amsmath,amsfonts}
\usepackage{algorithmic}
\usepackage{array}
\usepackage[caption=false,font=normalsize,labelfont=sf,textfont=sf]{subfig}
\usepackage{textcomp}
\usepackage{stfloats}
\usepackage{url}
\usepackage{verbatim}
\usepackage{graphicx}
\def\BibTeX{{\rm B\kern-.05em{\sc i\kern-.025em b}\kern-.08em
    T\kern-.1667em\lower.7ex\hbox{E}\kern-.125emX}}

\usepackage{makecell}
\usepackage{amsmath}
\usepackage{multirow}
\usepackage{tabularx}
\usepackage{booktabs}
\usepackage{adjustbox}
\usepackage{xcolor}
\usepackage[strings]{underscore}
\usepackage{bm}

\begin{document}

% \title{\myfont \textbf{Relative Localization for Resonant Beam Enabled UAV Swarm}} 

% \title{Relative Localization for Resonant Beam \protect\\ Enabled UAV Swarm}
% \title{Resonant Beam and Visual-Intertial Fusion for Positioning Optimization of UAV Swarm}
% \title{Resonant Beam-Enhanced Visual-Inertial Fusion Optimization for Multi-UAV Positioning}
\title{\huge Odometry-Aided Real-Time Mapping for Underwater Robots Using Forward-Looking Sonar}

% \author{\normalsize Siyuan Du$^{*12}$, Kanzhong Yao$^{*1}$, Youdong Wang$^1$, Yingqi Liu$^1$, Qingwen Liu$^2$, Qunhui Yang$^2$, Zhe Sun$^{\dagger1}$, Xuelong Li$^{\dagger1}$}

% \thanks{
% $^{1}$Institute of Artificial Intelligence (TeleAI), China Telecom, China.
% $^{2}$Tongji University, Shanghai, China.

%  }

\author{\normalsize Siyuan Du$^{*12}$, Kanzhong Yao$^{*1}$, Youdong Wang$^1$, Yingqi Liu$^1$, Qingwen Liu$^2$, Qunhui Yang$^2$, Zhe Sun$^{\dagger1}$, Xuelong Li$^{\dagger1}$% <-this % stops a space
\thanks{This work was done during an internship at TeleAI.}% <-this % stops a space
\thanks{$^{1}$Institute of Artificial Intelligence (TeleAI), China Telecom, China.}
\thanks{ $^{2}$Tongji University, Shanghai, China. }% 
\thanks{
E-mails: Siyuan Du (dueen1123@tongji.edu.cn),
Kanzhong Yao (yaokz1@chinatelecom.cn),
Youdong Wang (wangyd24@chinatelecom.cn),
Yingqi Liu (18553471556@163.com),
Qingwen Liu (qliu@tongji.edu.cn),
Qunhui Yang (yangqh@tongji.edu.cn),
Zhe Sun (sunzhe@nwpu.edu.cn),
and Xuelong Li (xuelong\_li@ieee.org).
}%
\thanks{$^{*}$Equal contribution.
$^{\dagger}$Corresponding authors.}
}

\markboth{IEEE }
{How to Use the IEEEtran \LaTeX \ Templates}

\maketitle
\begin{abstract}

Reliable perception is essential for underwater vehicles operating in complex environments, where light attenuation and scattering often degrade visibility and compromise optical sensing. Forward-looking sonar (FLS) offers an alternative by providing high-frame-rate acoustic imaging under poor optical conditions. However, real-time FLS mapping remains challenging due to unresolved target elevation, spatially non-uniform noise, and fragmented target boundaries, which hinder feature extraction and introduce geometric ambiguity during projection. To address these challenges, we propose a cascaded feature reconstruction pipeline combining fast Fourier transform (FFT)-based denoising, fast multiscale constant false alarm rate (MCFAR) detection, and gradient-adaptive boundary connection to extract geometric features from degraded sonar images with low latency. We integrate attitude-aware geometric projection with incremental occupancy accumulation to construct a depth-referenced 2.5D map for local mapping in confined underwater environments. The sonar's vertical position is referenced to an external sensor, while target elevation is assigned under an explicit geometric assumption rather than measured directly by FLS. Experiments in a 3 m $\times$ 5 m  pool demonstrate centimeter-scale planar mapping accuracy, with a root-mean-square error (RMSE) below 3 cm across three sequences and an average processing time of 42.4 ms per frame. 
% The source code and real-world datasets are publicly available.

\end{abstract}

% \begin{IEEEkeywords}
% Underwater Vehicles, Forward-Looking Sonar, Real-Time Mapping, Geometric Rectification.
% \end{IEEEkeywords}

%\IEEEpeerreviewmaketitle

\section{INTRODUCTION}
\label{sec:intro}

Underwater vehicles are widely used for missions including infrastructure inspection, hydrological data collection, and resource exploration~\cite{mccammon2017planning}. However, turbidity, suspended sediment, and darkness pose perceptual challenges that can impair traditional optical cameras~\cite{islam2020fast}\cite{sorensen2023commercial}\cite{sun2025water}. Although green-light underwater LiDAR provides 3D information, optical scattering and attenuation reduce its range in turbid waters, while high costs limit its adoption~\cite{11079880}. Acoustic perception is less affected by poor visibility, but Mechanically Scanned Imaging Sonars (MSIS) rely on physical rotation to emit single beams~\cite{hernandez2009probabilistic}. This leads to prolonged imaging cycles and motion distortion during navigation, limiting their ability to meet real-time requirements for agile obstacle avoidance. In contrast, FLS achieves high-frame-rate imaging via multi-beam arrays, supporting real-time perception in turbid and dark underwater environments~\cite{yan2024low}\cite{hurtos2015adaptive}\cite{sorensen2023commercial}.

\begin{figure}[h]

    \centering 
	\includegraphics[width=2.8in]{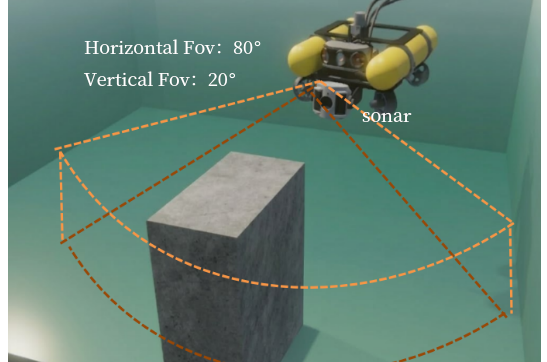}
	\caption{{Scenario of ROV underwater detection and obstacle perception.}}
	\label{fig:scene}
\end{figure}

Despite the advantages of FLS, autonomous underwater obstacle avoidance still faces algorithmic challenges. Many sonar-based strategies rely on reactive control driven by raw sonar images or end-to-end models~\cite{huang2025deep}~\cite{cao2022research}. These approaches provide limited global spatial understanding, restricting long-horizon path planning. In aerial robotics, agile autonomous navigation frameworks, such as Ego-Planner~\cite{zhou2020ego}, have demonstrated the benefits of real-time dense occupancy grid maps. This mapping-based navigation paradigm also holds promise for underwater applications. However, real-time FLS grid mapping faces non-uniform environmental noise, fragmented target edges, projection distortion due to missing vertical resolution, and multipath artifacts.

To address the aforementioned perception and mapping challenges, this paper proposes an odometry-aided real-time mapping system using FLS. Designed for confined underwater environments with degraded visibility, the system provides a reliable environmental representation for centimeter-scale local obstacle mapping and autonomous navigation.
It extracts robust geometric features from degraded sonar images and incorporates attitude information to correct projection distortions, thereby constructing a depth-referenced 2.5D occupancy grid map. Experiments are conducted in a (3$\times$5) m pool to evaluate short-range obstacle perception and local mapping, providing a foundation for local path planning and safe obstacle avoidance.

The main contributions of this paper are as follows:

\begin{itemize}

\item[1)] 
An efficient feature reconstruction pipeline for degraded sonar images that integrates frequency-domain denoising, fast MCFAR detection, and adaptive boundary connection to extract geometric features in real time under non-uniform noise.

\item[2)]
A probabilistic mapping model addressing FLS elevation ambiguity. By combining view-frustum distortion correction with a Log-Odds update mechanism, this approach physically suppresses projection ghosting, supporting 24 Hz real-time processing with centimeter-level mapping errors.

% \item[3)] A hardware implementation validated through multiple experiments in a pool with obstacles, with code and recorded datasets released to support further research\footnote{https://anonymous.4open.science/r/sonar_mapping-0C89}.

\item[3)] A hardware implementation validated through multiple trails in an experimental pool with obstacles.

\end{itemize}

\label{sec:RBSom}

\section{RELATED WORK}

\subsection{Mechanically Scanning Imaging Sonar Mapping}
Due to their robustness against poor lighting and turbidity, acoustic sensors are widely used for underwater perception. Early localization and SLAM frameworks mainly relied on MSIS~\cite{ribas2006slam}\cite{mallios2014scan}, using line-feature extraction and scan matching for structured environments and large-scale seafloor mapping. Although MSIS provides high-resolution imagery for static inspection, its sequential beam-scanning mechanism introduces long acquisition latency, causing motion distortion on highly maneuverable platforms (AUVs/ROVs)~\cite{ribas2008underwater}. This limitation makes MSIS unsuitable for agile obstacle avoidance and real-time dynamic mapping.

\subsection{Multimodal Mapping: Fusion of Vision and FLS}

To address the real-time limitations of MSIS, high-frame-rate multibeam FLS has been widely adopted. However, FLS projects three-dimensional acoustic returns onto a two-dimensional image, leaving target elevation unresolved. To mitigate this ambiguity, multimodal mapping approaches integrate FLS with complementary visual sensors, such as monocular or stereo cameras.

Early fusion methods aligned independently reconstructed optical and acoustic 3D models, but their computational cost limited real-time operation~\cite{kim20193}. Tightly coupled systems, such as SVin2~\cite{rahman2022svin2} and the method of Cardaillac et al.~\cite{cardaillac2023camera}, improved efficiency using visual observations for odometry, yet produced sparse landmark maps with limited utility for ROV obstacle avoidance and path planning~\cite{wang2023real}.

Recent work has shifted toward cross-modal dense 3D mapping. Collado-Gonzalez et al.~\cite{collado2025opti} associated visual regions of interest with sonar echo clusters for geometric reconstruction. Learning-based methods include SonarSweep~\cite{chen2025sonarsweep}, which uses plane-sweeping cost volumes for dense depth estimation, and OA-Stereo~\cite{cao2025oa}, which refines disparity through self-supervised opti-acoustic matching. Gutnik et al.~\cite{gutnik2024enhancing} also explored self-supervised sensor fusion for AUV 3D obstacle avoidance.

Despite these advances, opti-acoustic mapping still depends on usable visual information. In highly turbid waters, optical backscatter severely reduces image contrast and obscures texture~\cite{hidalgo2015review}, potentially compromising visual association and depth estimation. This limitation motivates sonar-based mapping under poor visibility.

\subsection{FLS-Only Perception and Mapping}

To achieve reliable perception in optically denied environments, researchers have pursued FLS-only mapping. However, without multimodal assistance, balancing real-time performance and mapping fidelity remains a formidable challenge.

% At the perception front-end, early CFAR-based methods~\cite{wang2022virtual} extract acoustic features that are highly susceptible to fragmentation and misassociation under severe FLS noise. To resolve FLS elevation ambiguity, methods relying on specific motion constraints—such as multi-view space carving~\cite{guerneve2018three} or 90-degree rolling~\cite{park20233d}—have emerged, but these severely restrict the vehicle's operational efficiency. Recently, Zhi et al.~\cite{zhi2025oscillatory} introduced an oscillating FLS mechanism for 3D coverage. Yet, their reliance on naive thresholding for real-time noise removal destroys the topological continuity of target edges, failing to meet high-fidelity reconstruction requirements.

At the perception front-end, early CFAR-based methods~\cite{wang2022virtual} extract acoustic features susceptible to fragmentation and misassociation under severe FLS noise. To resolve FLS elevation ambiguity, methods using specific motion constraints, such as multi-view space carving~\cite{guerneve2018three} or 90-degree rolling~\cite{park20233d}, have emerged, but these restrict operational efficiency. Recently, Zhi et al.~\cite{zhi2025oscillatory} introduced an oscillating FLS mechanism for 3D coverage. However, their reliance on simple thresholding for real-time denoising can disrupt target-edge continuity, limiting reconstruction fidelity.

% Regarding back-end map representation, high-precision volumetric albedo models~\cite{westman2020volumetric} are computationally prohibitive for online execution. Conversely, real-time grid mapping methods~\cite{cheng2022underwater}~\cite{mcconnell2025above} often idealize FLS images as 2D rays to prioritize efficiency. This oversimplification ignores the projection distortions caused by dynamic ROV pitch motions, resulting in severe ghosting artifacts and positional deviations around complex 3D structures.

Regarding back-end map representation, high-precision volumetric albedo models~\cite{westman2020volumetric} are computationally prohibitive for online execution. Conversely, real-time grid mapping methods~\cite{cheng2022underwater}~\cite{mcconnell2025above} often idealize FLS images as 2D rays for efficiency. This simplification ignores projection distortions caused by ROV pitch motions, resulting in ghosting artifacts and positional deviations around complex 3D structures.

To break the inherent trade-off between real-time performance and reconstruction accuracy, our proposed framework introduces a cascaded feature reconstruction pipeline and a frustum geometric correction mechanism. This approach achieves centimeter-level precision while maintaining a 24 Hz real-time processing capability.

% To address the trade-off between real-time performance and reconstruction accuracy, our framework introduces a cascaded feature reconstruction pipeline and a frustum geometric correction mechanism, achieving centimeter-level precision at a processing rate of 24 Hz.

\begin{figure*}[h]

    \centering 
	\includegraphics[width=5.7in]{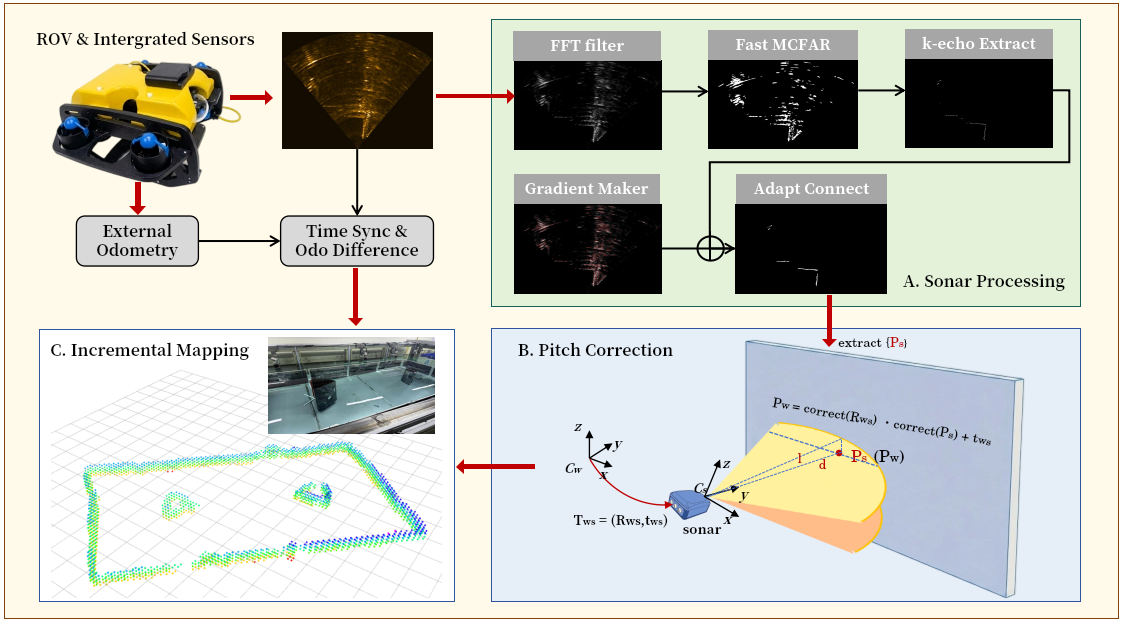}
	\caption{{Overview of the proposed sonar mapping pipeline: Temporal alignment and interpolation of multi-sensor data, (A) Sonar image processing pipeline, (B) Frustum-based coordinate rectification, and (C) Incremental probabilistic mapping performance.}}
	\label{fig:mapping pipeline}
\end{figure*}

\section{Methodology}

\subsection{Sonar Image Processing}

\subsubsection{Frequency-Domain Filtering}

Sonar images often suffer from concentric arc artifacts caused by hardware crosstalk or range-dependent background noise, which manifest as horizontal stripes in polar coordinates. To mitigate this, we introduce a 2D-FFT-based frequency-domain filtering preprocessing method. First, the polar image $I_{pad}(x,y)$ is converted into the frequency domain:
\begin{equation}
\begin{aligned}
\mathcal{F}(u, v) = \sum_{x=0}^{M-1} \sum_{y=0}^{N-1} I_{pad}(x, y) e^{-j 2\pi \left(\frac{ux}{M} + \frac{vy}{N}\right)},
\end{aligned}
\end{equation}
where  $M$ and $N$ denote the image dimensions, while $u$ and $v$ represent the vertical and horizontal spatial frequencies. Since horizontal stripes lack horizontal intensity variations, their energy is concentrated in the extremely low-frequency region. To accurately filter out this interference, a frequency mask is designed as follows ($W$ denotes the cutoff bandwidth):
% \begin{equation}
% \begin{aligned}
% H(u, v) = 
% \begin{cases}
% 0, & v \in [0, W-1] \cup [N-W, N-1] \\
% 1, & \text{otherwise}
% \end{cases} .
% \end{aligned}
% \end{equation}
\begin{equation}
H(u,v)=
\begin{cases}
0, & v\in[0,W-1]\cup[N-W+1,N-1],\\
1, & \text{otherwise},
\end{cases}
\end{equation}

Direct frequency truncation can induce the Gibbs phenomenon, causing negative artifacts in the reconstructed image. Thus, an inverse Fourier transform with a non-negativity constraint is applied to the filtered spectrum to compute the final reconstructed image, $I_{out}(x, y)$:
\begin{equation}
\begin{aligned}
I_{out}(x, y) = \max \left( 0, \, \Re \left\{ \mathcal{F}^{-1} \left\{ \mathcal{F}(u, v) H(u, v) \right\} \right\} \right),
\end{aligned}
\end{equation}
where $\Re\{\cdot\}$ denotes the real part of a complex number, $\mathcal{F}^{-1}\{\cdot\}$ represents the inverse Fourier transform.

\subsubsection{Fast Multiscale Truncated CFAR}

Due to severe reverberation variations in sonar images, traditional fixed thresholds fail to balance detection and false alarm rates. We introduce the CFAR. Assuming exponentially distributed noise, the decision threshold $T = \alpha \hat{\sigma}$ is computed using the noise expectation $\hat{\sigma}$ from $N$ reference cells. To maintain a preset false alarm probability ($P_{fa}$), the exact threshold coefficient $\alpha_s$ is given by:
\begin{equation}
\begin{aligned}
\alpha_s = N \left( P_{fa}^{-\frac{1}{N}} - 1 \right),
\end{aligned}
\end{equation}

Since MCFAR is computationally prohibitive for real-time ROV obstacle avoidance, we propose a Fast Multiscale Truncated CFAR using prefix sum acceleration. For single-beam data $I(r)$ ($r \in [0, M-1]$), a 1D prefix sum is constructed as follows:
\begin{equation}
\begin{aligned}
P(k) = \sum_{i=0}^{k-1} I(i), \quad P(0) = 0,
\end{aligned}
\end{equation}

At any scale $s$, with an outer training window radius $R_{t, s}$ and an inner guard window radius $R_g$, the total energy of the local training window $E_{train, s}$ can be directly calculated using the prefix sum and boundary truncation functions:
\begin{equation}
\begin{aligned}
E_{train, s} = &\big[P(r_{out\_end} + 1) - P(r_{out\_start})\big]  \\
&- \big[P(r_{in\_end} + 1) - P(r_{in\_start})\big],
\end{aligned}
\end{equation}
where $r_c$ is the index of the cell under test; $r_{out\_start} = \max(0, r_c - R_{t, s})$ and $r_{out\_end} = \min(M-1, r_c + R_{t, s})$ denote the boundary-truncated start and end indices of the outer training window, respectively; and $r_{in\_start} = \max(0, r_c - R_g)$ and $r_{in\_end} = \min(M-1, r_c + R_g)$ represent the truncated start and end indices of the inner guard window.

The local noise expectation is given by $\hat{\sigma}_s = E_{train, s} / N_{train, s}$. To prevent thermal noise amplification in deep water or acoustic shadows where $\hat{\sigma}_s \to 0$, a sensitivity lower bound $T_{min}$ is introduced to form the truncated threshold $T_s$:
\begin{equation}
\begin{aligned}
T_s = \max \big(\alpha_s \hat{\sigma}_s, \ T_{min} \big),
\end{aligned}
\end{equation}

To enhance robustness against complex reverberation, a Majority Voting strategy is applied. The final decision $D_{final}(r_c)$ is true only when the threshold is exceeded in most scales:
\begin{equation}
\begin{aligned}
D_{final}(r_c) = \begin{cases} 1, & \sum_{s=1}^S \mathbb{I}\big(I(r_c) > T_s\big) \ge \lfloor S/2 \rfloor + 1 \\ 0, & \text{otherwise} \end{cases},
\end{aligned}
\end{equation}
where $S$ denotes the total number of scales, $I(r_c)$ is the intensity value of the cell at position $r_c$, $T_s$ represents the detection threshold at the $s$-th scale, and $\mathbb{I}(\cdot)$ is the indicator function.

This mechanism efficiently suppresses isolated noise, yielding clean, high-fidelity data for subsequent 3D point cloud reconstruction.

\subsubsection{Top-$k$ Echo Extraction and Adaptive Connection}

Sonar targets typically produce strong front-edge echoes followed by noisy tails. To suppress these tails, we scan each beam from near to far and retain only the first $k$ detected target pixels, preserving the target's front edge.

Since these edges are often fragmented, standard morphological closing may excessively thicken thin targets. We therefore use an adaptive directional connection method based on the structure tensor. The local edge direction $\phi(x,y)$ is estimated from its components ($J_{xx}, J_{yy}, J_{xy}$):
% \begin{equation}
% \begin{aligned}
% \phi(x,y) = \frac{1}{2} \arctan \left( \frac{2J_{xy}}{J_{xx} - J_{yy}} \right) + \frac{\pi}{2}.
% \end{aligned}
% \end{equation}
\begin{equation}
\phi(x,y)=
\frac{1}{2}\operatorname{atan2}
\left(2J_{xy},J_{xx}-J_{yy}\right)
+\frac{\pi}{2},
\end{equation}

The orientation field is discretized into angular subsets, each assigned a directional elliptical structuring element. To reduce computation, morphological closing is restricted to minimal bounding boxes within dynamic ROIs, and the results are mapped back to the full image. This approach bridges edge gaps while limiting target thickening and maintaining real-time performance.

% We then divide these directions into several angle groups. For broken points in each group, we connect them using a long elliptical kernel that matches their specific direction. Finally, we combine the results from all angles. This method successfully repairs the gaps while keeping the exact width and shape of the targets.

\subsection{3D Distortion Compensation via Frustum-Edge Model}

FLS projects 3D acoustic echoes onto a 2D plane, losing elevation resolution. Conventional 3D reconstruction assumes $\phi=0$, which causes severe point cloud distortion during pitching maneuvers. To ensure accurate mapping, this paper proposes an attitude compensation method based on frustum-edge approximation.

\begin{figure}[h]

    \centering 
	\includegraphics[width=2.5in]{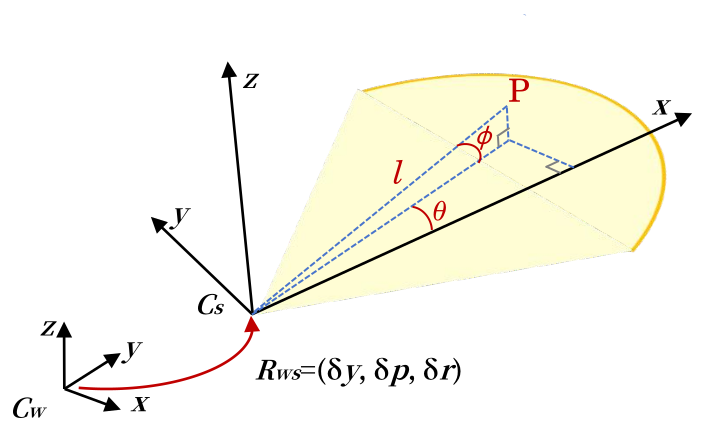}
	\caption{{Geometric model of the FLS, illustrating the transformation from the world frame $C_w$ to the sonar frame $C_s$, and the representation of a spatial point $P$ within the acoustic frustum.}}
	\label{fig:Geometric model}
\end{figure}

First, invalid echoes caused by strong sea-surface reflections during high-pitch maneuvers must be eliminated. Under a two-dimensional side-view approximation with a flat water surface, the upward-facing upper frustum boundary intersects the surface at the slant range $R_{surf}$:
\begin{equation}
\begin{aligned}
R_{surf} = \frac{H}{\sin\left(\delta p + \frac{\Phi_v}{2}\right)},
\end{aligned}
\end{equation}
where $H$ denotes the real-time depth of the sonar, $\delta p$ is the sonar pitch angle, and $\Phi_v$ is the vertical field of view (FOV).

The system identifies pixels with range $l > R_{surf}$ as sea-surface multipath reverberation and filters them out, ensuring the physical consistency of the projected data. For the remaining valid echoes, a 3D point $P_s = [x_s, y_s, z_s]^T$ in the sonar coordinate frame can be theoretically represented in the standard spherical coordinate system as:
\begin{equation}
\begin{aligned}
P_s = \begin{bmatrix} l \cos\phi \cos\theta \\ l \cos\phi \sin\theta \\ l \sin\phi \end{bmatrix},
\end{aligned}
\end{equation}
where $l$ denotes the range, $\theta$ is the horizontal azimuth angle, and $\phi$ represents the vertical elevation angle as Fig.~\ref{fig:Geometric model}. Since the exact elevation angle $\phi$ of the echo cannot be directly obtained, conventional methods typically assume $\phi = 0$ for 3D projection. However, this leads to significant point cloud distortion when the vehicle undergoes substantial pitching motion.

\begin{figure}[h]
    \centering 
	\includegraphics[width=3.5in]{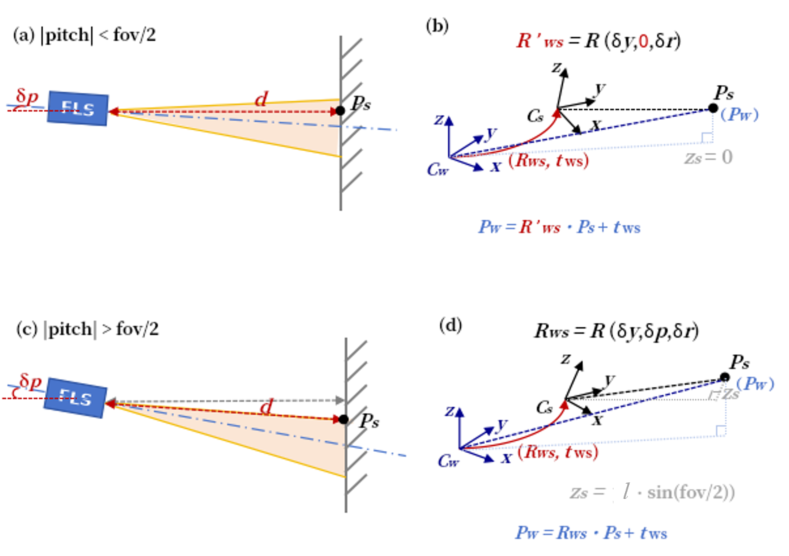}
	\caption{{Side view of the effect of vehicle pitch on FLS distance measurement. Here, $P$ denotes the reflection point generating the shortest echo distance $d$ within the frustum, and $\delta p$ represents the current pitch angle of the sonar.}}
	\label{fig:Side view}
\end{figure}

To accurately map the local sonar point $P_s$ to the world coordinate system, we employ the following transformation, and $\{R_{ws}, t_{ws}\}$ 
denotes the rotation from the sonar frame to the world frame:
\begin{equation}
\begin{aligned}
P_w = R_{ws}P_s + t_{ws},
\end{aligned}
\end{equation}

We adopt a piecewise projection model based on the sonar pitch angle. For small pitch angles, an empirical pitch-suppression strategy is applied. For larger pitch angles, an assumed frustum-boundary elevation is substituted into the spherical-coordinate expression without approximating the trigonometric terms. The three cases are described below and illustrated in Fig.~\ref{fig:Side view}.

\subsubsection{{Small Pitch Angle State $|\delta {p}| \le \Phi_v / 2$}}

When the vehicle's pitch angle remains within the vertical FOV, the central region of the sonar frustum covers the horizontal extension plane in the world coordinate system. In this case, the detected targets at the leading edge can be assumed to be captured by the central beam (i.e., the elevation angle is approximated as $\phi \approx 0$). Consequently, the 3D coordinates $P_s$ in the sonar frame are represented as:
\begin{equation}
\begin{aligned}
P_s = \begin{bmatrix} l \cos\theta \\ l \sin\theta \\ 0 \end{bmatrix},
\end{aligned}
\end{equation}

% During the transformation to the world coordinate system, to eliminate the projection distortion caused by minor vehicle pitching, the pitch component in the rotation matrix is forcibly set to zero. Thus, the rotation matrix $R_{ws}$ is calculated using the modified attitude angles:
To reduce the influence of small pitch variations on the map, the pitch component of the projection rotation is set to zero while yaw and roll are retained:
\begin{equation}
\begin{aligned}
R_{ws} \leftarrow \mathcal{R}(\delta y, 0, \delta {r}),
\end{aligned}
\end{equation}
where $\mathcal{R}(\cdot)$ denotes the mapping from Euler angles (yaw, pitch, roll) to the corresponding rotation matrix in $SO(3)$, $\delta y$ and $\delta r$ denote the sonar yaw and roll angle, respectively.

\subsubsection{{Large Positive Pitch State $\delta {p} > \Phi_v / 2$}}

When the vehicle’s upward pitch exceeds half the vertical FOV, the earliest return is assumed to originate from the lower boundary of the sonar frustum. Substituting the lower frustum boundary angle into the spherical-coordinate expression yields the estimated echo-point coordinates in the sonar frame:
% \begin{equation}
% \begin{aligned}
% P_s = \begin{bmatrix} l  \cos\theta \\ l \sin\theta \\ -l \cos\theta \sin\left(\frac{\Phi_v}{2}\right) \end{bmatrix}.
% \end{aligned}
% \end{equation}
\begin{equation}
\begin{aligned}
P_s = \begin{bmatrix} l \cos\left(\frac{\Phi_v}{2}\right)  \cos\theta \\ l  \cos\left(\frac{\Phi_v}{2}\right) \sin\theta \\ -l  \sin\left(\frac{\Phi_v}{2}\right) \end{bmatrix},
\end{aligned}
\end{equation}
When transforming to the world coordinate system, the true 3D attitude $R_{ws}$ of the sonar must be retained for rigorous mapping:
\begin{equation}
\begin{aligned}
R_{ws} = \mathcal{R}(\delta {y}, \delta {p}, \delta {r}),
\end{aligned}
\end{equation}

\begin{figure*}[!t]

    \centering 
	\includegraphics[width=6.0in]{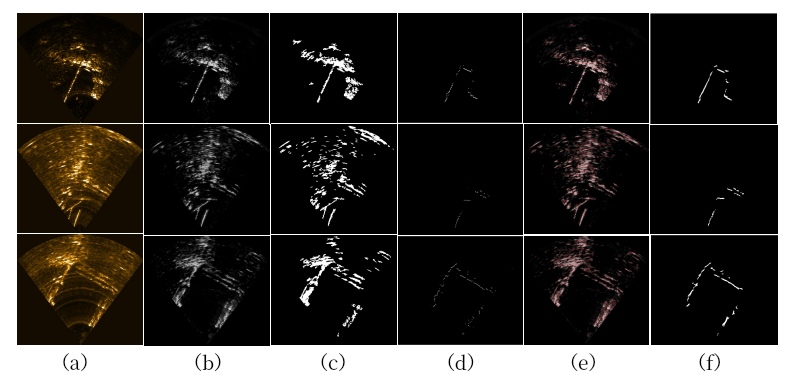}
	\caption{{The results of sonar image processing pipeline: (a) Raw image, (b) FFT-based denoising, (c) Fast-MCFAR and dual-threshold filtering, (d) Top-k echo extraction, (e) Target gradient estimation derived from (b), and (f) Adaptive connection based on (d) and (e).}}
	\label{fig:sonar process}
\end{figure*}

\subsubsection{{Large Negative Pitch State $\delta {p} < -\Phi_v / 2$}}

Similarly, when the vehicle’s downward pitch exceeds half the vertical FOV, the earliest return is assumed to originate from the upper boundary of the sonar frustum. 
Substituting the upper frustum boundary angle into the spherical-coordinate expression yields the estimated echo-point coordinates in the sonar frame:
% \begin{equation}
% \begin{aligned}
% P_s = \begin{bmatrix} l \cos\theta \\ l \sin\theta \\ l \cos\theta \sin\left(\frac{\Phi_v}{2}\right) \end{bmatrix}.
% \end{aligned}
% \end{equation}
\begin{equation}
\begin{aligned}
P_s = \begin{bmatrix} l \cos\left(\frac{\Phi_v}{2}\right) \cos\theta \\ l \cos\left(\frac{\Phi_v}{2}\right) \sin\theta
\\ l  \sin\left(\frac{\Phi_v}{2}\right) \end{bmatrix},
\end{aligned}
\end{equation}
The coordinate transformation employs the true sonar attitude $R_{ws}$:
\begin{equation}
\begin{aligned}
R_{ws} = \mathcal{R}(\delta {y}, \delta {p}, \delta {r}),
\end{aligned}
\end{equation}

In conclusion, the proposed method effectively addresses the sonar's lack of elevation resolution, significantly suppressing 3D point cloud distortion during severe maneuvers. Admittedly, this model relies on a Structured Environment Assumption (assuming flat seabeds or regular walls) and may introduce local elevation errors in highly complex terrains. However, compared to the global distortion of traditional 2D assumptions, our method drastically improves geometric consistency with minimal computational overhead.

\subsection{Incremental Log-Odds Occupancy Mapping}

To suppress sporadic sonar noise, we maintain a probabilistic voxel map through multi-frame hit accumulation, using spatial hashing to allocate memory only for voxels containing valid echo endpoints. Let $L_t(n)$ denote the occupancy log-odds of voxel $n$ at time $t$. Following Bayesian odds updating, hit evidence is accumulated additively via a log-likelihood ratio with a fixed factor $\lambda_{\mathrm{hit}} = p_{\mathrm{hit}} / (1 - p_{\mathrm{hit}})$, where $p_{\mathrm{hit}}$ is a prescribed weighting parameter rather than a calibrated detection probability. The update rule is defined as:
\begin{equation}
L_t(n) = \begin{cases}
\min\!\left[L_{t-1}(n) + \ln\lambda_{\mathrm{hit}},\, L_{\max}\right], & n \in \mathcal{H}_t, \\[4pt]
L_{t-1}(n), & n \notin \mathcal{H}_t.
\end{cases}
\end{equation}
% \begin{equation}
% \bm{L}_t(n)=
% \begin{cases}
% \min\!\left[\bm{L}_{t-1}(n)+\ln\lambda_{\mathrm{hit}},\,L_{\max}\right],
% & n\in\mathcal{H}_t,\\[4pt]
% \bm{L}_{t-1}(n),
% & n\notin\mathcal{H}_t.
% \end{cases}
% \end{equation}
where $\mathcal{H}_t$ is the set of voxels with valid echo endpoints in frame $t$. Newly allocated voxels are initialized to $L_0 = \operatorname{logit}(p_{\min})$, where $\operatorname{logit}(p) = \ln[p/(1-p)]$. To prevent confidence inflation, each voxel is updated at most once per frame, treating hit evidence across frames as approximately independent. Voxels without hits retain their prior occupancy estimates.
Only voxels satisfying $L_t(n) > L_{\mathrm{occ}}$ are extracted, where $L_{\mathrm{occ}} = \operatorname{logit}(p_{\mathrm{occ}})$. Requiring hits across multiple frames effectively suppresses sporadic noise, while the upper bound $L_{\max} = \operatorname{logit}(p_{\max})$ caps accumulated confidence. Map extraction traverses only allocated voxels, avoiding a full volume scan.

\section{EXPERIMENTS}

\subsection{Hardware Overview}

The experimental platform is an ROV equipped with an SV1210 FLS (1.8 MHz, $80^\circ$ horizontal and $20^\circ$ vertical FOV). External pose estimates may be provided by navigation systems incorporating DVL, inertial, or acoustic measurements, or by LiDAR-based odometry when above-water mounting is feasible. In this work, a Livox Mid-360 LiDAR is mounted above the water on a custom support, and the sonar attitude and depth are derived from LiDAR odometry through extrinsic calibration. A separate scan of the drained pool is used as the reference map for evaluation. Onboard synchronization and algorithm execution are performed on a Holybro Pixhawk Jetson Baseboard~\cite{HolybroBaseboard2024}. Multiple sequences were collected in the pool to evaluate mapping accuracy and runtime performance.

\subsection{Performance of sonar image processing}
Fig.~\ref{fig:sonar process} illustrates the proposed FLS image processing pipeline. First, an FFT filter suppresses inherent transverse wave artifacts in the raw Cartesian image (Fig.~\ref{fig:sonar process}a, b). Next, the proposed Fast MCFAR algorithm and connected-component removal extract a clean binary target mask (Fig.~\ref{fig:sonar process}c), while the 2D gradient field is computed concurrently (Fig.~\ref{fig:sonar process}e). Finally, guided by the gradient field, an Adaptive Connection (AC) algorithm bridges edge gaps from the initial front echo to extract a continuous and complete physical contour (Fig.~\ref{fig:sonar process}f).

Table I evaluates the pipeline's real-time performance. Note that the total "Sonar Process" time exceeds the sum of its core sub-steps due to additional processing steps, such as polar-to-Cartesian coordinate transformations and morphological filtering.

\subsection{Evaluation of Mapping Metrics}

\begin{figure*}[h]

    \centering 
	\includegraphics[width=7.0in]{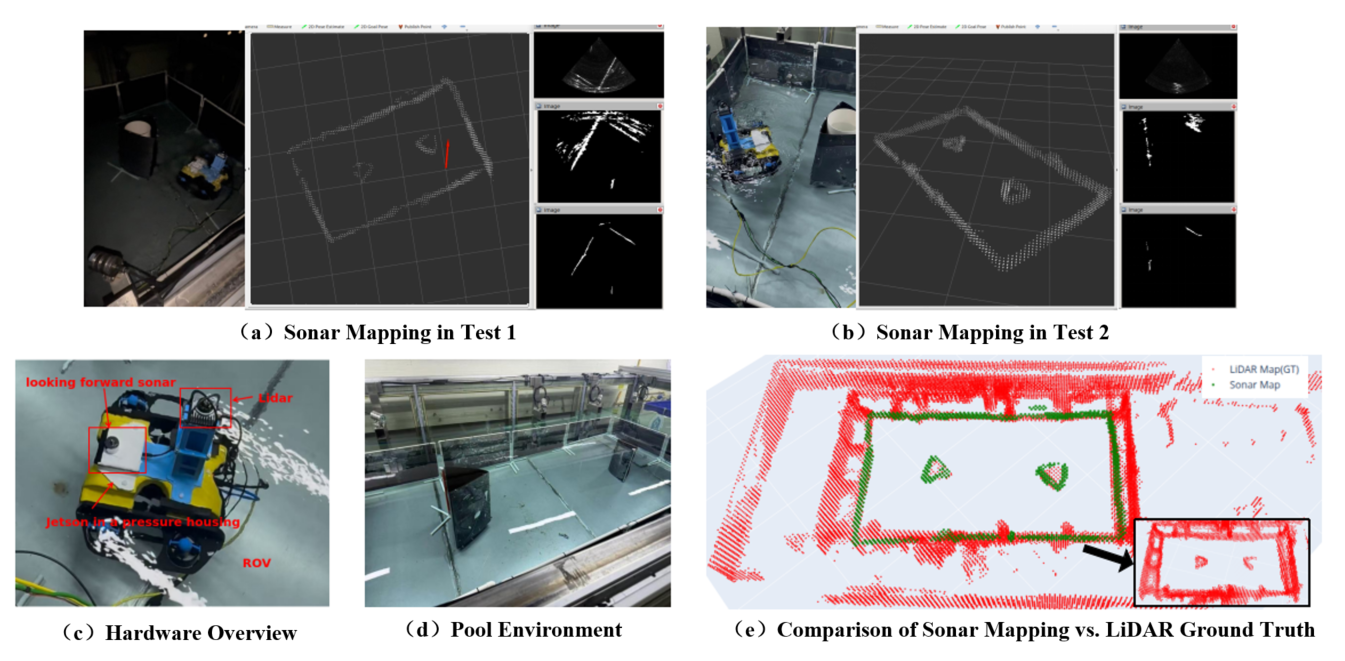}
	\caption{{Experimental results. (a) Real-time mapping in test 1 (14_24 dataset). (b) Real-time mapping in test 2 (10_35 dataset). (c) ROV equipped with both an underwater sonar and a surface LiDAR, (d) A 3m x 5m indoor testing tank with obstacles. (e) Mapping results of the 10_35 dataset. The red point cloud is the LiDAR ground truth captured in the drained pool, while the green point cloud is the underwater sonar map.}}
	\label{fig: exp scene}
\end{figure*}

Fig.~\ref{fig: exp scene} presents the experimental setup and the corresponding 3D mapping results. As shown in Fig.~\ref{fig: exp scene}(d), experiments were conducted in a $3\,\mathrm{m}\times5\,\mathrm{m}$ experimental pool containing two triangular obstacles with distinctive sonar features. Multiple trials were performed to evaluate system robustness. Fig.~\ref{fig: exp scene}(e) compares the sonar map (green) with the LiDAR-derived reference map (red).

\begin{figure}[h]

    \centering 
	\includegraphics[width=3.5in]{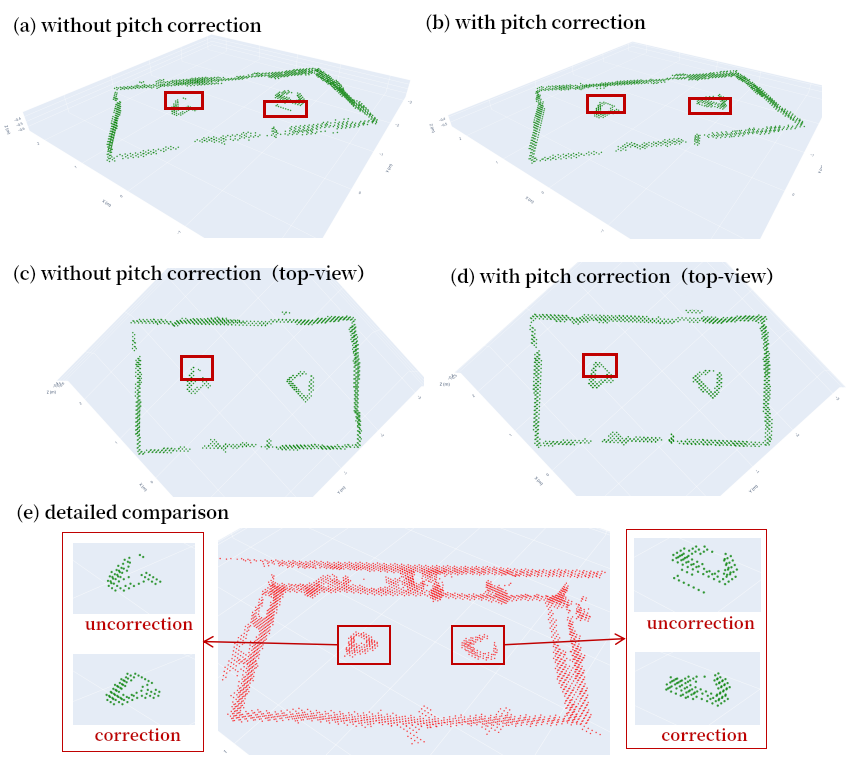}
	\caption{{Sonar mapping comparison with and without pitch rectification in 10_35 dataset.}}
	\label{fig:pitch comparison}
\end{figure}

\begin{table}[htbp]
\centering
\caption{Performance Metrics}
\begin{tabular}{lcc}
\toprule
 & Mean time (ms) & STD (ms) \\
\midrule
\textbf{Sonar Process}   & \textbf{41.86} & \textbf{4.92} \\
  \quad 1) FFT & 10.94 & 1.63 \\
  \quad 2) MCFAR$^\dagger$ & 74.53 & 1.47 \\
  \quad 2) MCFAR Fast  & 2.83 & 0.52 \\
  \quad 3) AC$^\dagger$  & 24.16 & 1.90 \\
  \quad 3) AC Fast  & 17.38 & 2.41 \\
 % \bottomrule
\midrule
Mapping  & 165.02 & 4.88 \\
\textbf{Mapping Fast} & \textbf{0.54} & \textbf{0.25} \\
\midrule
\textbf{Total Time}  & \textbf{42.4} & \textbf{5.03} \\
\bottomrule
\end{tabular}
\end{table}

\subsubsection{Accuracy Analysis}
Table II compares the sonar mapping accuracy across different test sequences. During underwater navigation, the ROV inevitably experiences pitch fluctuations, preventing a strictly stable attitude. 
% Interestingly, these dynamic fluctuations enable the resulting grid map to capture a depth distribution of approximately 0.2 m to 0.4 m along the Z-axis. 
The reconstructed points span approximately 0.2--0.4 m
along the Z-axis under the adopted projection model.
Consequently, we evaluate both 2D and 3D mapping errors. The 2D error is calculated by projecting the point cloud onto the XY plane and computing the absolute planar distance to the ground truth, while the 3D error is determined by the nearest-neighbor distance in 3D space.

Multiple sequences were collected to evaluate the proposed mapping method. Due to the sonar's narrow vertical FOV, pitch oscillations in structured environments can shift target echoes outside the effective detection range, causing ghosting artifacts without correction. As demonstrated by the quantitative results in Fig.~\ref{fig:pitch comparison} and Table~II, the proposed pitch compensation mechanism effectively suppresses ghosting interference and improves mapping accuracy, achieving a 2D RMSE below 3 cm.

\begin{table}[htbp]
\centering
\caption{Mapping errors with and without pitch correction.}
\begin{tabular}{ccccc}
\toprule
\textbf{Dataset} & \textbf{Dim} & \textbf{Type} & \textbf{AVE (cm)} & \textbf{RMSE (cm)} \\
\midrule
\multirow{4}{*}{10\_35} & \multirow{2}{*}{2D} & Uncorrected & 3.14 & 4.92 \\
                        &                     & Corrected   & 1.36 & 2.94 \\
\cmidrule{2-5}
                        & \multirow{2}{*}{3D} & Uncorrected & 7.19 & 8.69 \\
                        &                     & Corrected   & 4.13 & 5.67 \\
\midrule
\multirow{4}{*}{14\_24} & \multirow{2}{*}{2D} & Uncorrected & 1.75 & 3.13 \\
                        &                     & Corrected   & 0.62 & 1.79 \\
\cmidrule{2-5}
                        & \multirow{2}{*}{3D} & Uncorrected & 3.36 & 4.71 \\
                        &                     & Corrected   & 2.10 & 3.78 \\
\midrule
\multirow{4}{*}{15\_09} & \multirow{2}{*}{2D} & Uncorrected & 2.09 & 4.23 \\
                        &                     & Corrected   & 0.77 & 2.83 \\
\cmidrule{2-5}
                        & \multirow{2}{*}{3D} & Uncorrected & 5.49 & 7.17 \\
                        &                     & Corrected   & 2.06 & 4.04 \\
\bottomrule
\end{tabular}
\end{table}

\subsubsection{Runtime Analysis}

% System runtime was evaluated on an NVIDIA Jetson Orin NX (16GB) integrated with a Holybro Pixhawk Jetson Baseboard, using data from the 3 m $\times$ 5 m pool experiments. Incremental mapping updates only voxels affected by the current frame, reducing the per-frame map update time from 165 ms for the tested dense occupancy grid implementation to under 1 ms. Combined with optimized sonar image processing, the pipeline runs at an average of 42.4 ms per frame, or approximately 24 Hz, supporting real-time local mapping for small underwater robots.

Algorithm runtime was evaluated on an NVIDIA Jetson Orin NX (16GB) using data collected from the $3\mathrm{m}\times5\mathrm{m}$ pool experiments. By updating only voxels affected by the current frame, incremental mapping reduces the per-frame map update time from 165 ms for the tested dense grid implementation to under 1 ms. With optimized sonar image processing, the pipeline averages 42.4 ms per frame (approximately 24 Hz), supporting real-time local mapping. These timings reflect only algorithm execution on the NX, excluding inter-module communication and other system-level delays.

% To validate the superiority of the proposed mapping algorithm, we compare it with a few representative state-of-the-art works on real-time FLS mapping.
\subsection{Comparison with Existing Methods}
% To contextualize the proposed method, we summarize published results from related FLS mapping studies.
% Since their implementations are not publicly available, we directly adopt the experimental setups and results from the original papers for comparison.

We summarize published results from related FLS mapping studies for comparison. Since their implementations remain unavailable, including Ref~\cite{mcconnell2025above}, whose code was described as open source but remained unavailable at the time of our submission, we use the experimental setups and results reported in the original papers.

\begin{table}[htbp]
\centering
\caption{Comparison with Existing Methods} % 可以修改为你的标题
\begin{tabular}{lccc}
\toprule
 & Dimension (m) & RMSE (m) & Runtime (ms) \\
\midrule
Ref~\cite{zhi2025oscillatory}            & 1 $\times$ 2    & 0.139 &  --     \\    
                  & 1 $\times$ 6    & 0.203 &  --     \\    
Ref~\cite{mcconnell2025above}            & 20 $\times$ 50      & 3.37      & 887.8 \\

\textbf{Proposed} & \textbf{3 $\times$ 5} & \textbf{0.025}   & \textbf{42.4 } \\
\bottomrule
\end{tabular}
\end{table}

Ref.~\cite{zhi2025oscillatory} constructed an FLS grid map of lake bridge piers using direct thresholding for outlier rejection, reporting mapping errors of 0.139 m for 2-m spacing and 0.122 m / 0.203 m for 6-m spacing. Ref.~\cite{mcconnell2025above} performed FLS mapping on a Kingfisher USV using above-water LiDAR odometry. With relatively stable platform pitch, no attitude correction was applied. The reported RMSEs ranged from 2 to 15 m in bridge and harbor environments spanning 20--200 m, mainly due to accumulated odometry drift. In a different application setting, our method targets short-range obstacle avoidance and navigation for small underwater robots. It uses pitch compensation to suppress map ghosting caused by ROV attitude variations, achieving a planar mapping RMSE below 3 cm in a 3 m $\times$ 5 m pool and demonstrating its local mapping capability in confined environments.

Regarding computational cost, the mean runtimes reported in Ref.~\cite{mcconnell2025above} for point cloud registration, rectangle compression, and outlier rejection sum to 887.8 ms, whereas our 42.4 ms covers sonar image processing and incremental mapping. Although the timing scopes differ, our processing rate of approximately 24 Hz supports real-time local mapping for small underwater robots.

\section{Conclusions and Discussion}

This paper presented a high-precision, real-time incremental mapping system for FLS. By accounting for projection ambiguity associated with pitch variations, the proposed frustum-edge model reduces ghosting artifacts and improves geometric consistency during dynamic maneuvers. Experiments demonstrate a planar mapping RMSE below 3 cm across three sequences, with an average processing time of 42.4 ms per frame (approximately 24 Hz).

The current method has three limitations: (1) The correction model assumes structured environments, such as flat seabeds or vertical walls, and may introduce local elevation errors in highly irregular terrain. Nevertheless, this assumption is suited to typical 2.5D underwater operations, where depth regulation allows navigation and obstacle avoidance to focus primarily on the horizontal plane. (2) Multi-sensor data are aligned using timestamps without explicit compensation for communication delays between sensors and processing modules. (3) The map accumulates only hit evidence and lacks a mechanism to clear outdated occupancy, limiting its adaptability to dynamic environments.

Future work will focus on two directions: (1) Integrate the mapping framework with trajectory generation and incorporate communication latency compensation to support autonomous underwater navigation and obstacle avoidance; (2) Investigate visibility-aware free-space updates and temporal decay to clear outdated occupancy and improve adaptation to dynamic environments.

\label{sec:con}



 
\bibliographystyle{IEEETran}
\small 
%\bibliographystyle{IEEEtr}
\bibliography{mybib}
%\begin{thebibliography}{1}
%\end{thebibliography}


%\end{IEEEbiography}

%\newpage



%\vfill



\end{document}